\documentclass[pdflatex,sn-mathphys-num,twocolumn]{sn-jnl}

\usepackage[numbers,sort&compress]{natbib}
\usepackage{amsmath,amssymb}
\usepackage{graphicx}
\usepackage{booktabs}
\usepackage{siunitx}
\usepackage{xcolor}
\usepackage{soul}
\usepackage{placeins}
\usepackage[colorinlistoftodos]{todonotes}
\usepackage{acro}

\DeclareAcronym{AAS}{
  short = AAS,
  long  = Asset Administration Shell,
}

\DeclareAcronym{AI}{
  short = AI,
  long  = Artificial Intelligence,
}

\DeclareAcronym{CPS}{
  short = CPS,
  long  = Cyber-Physical System,
  short-plural-form = CPSs,
  long-plural-form  = Cyber-Physical Systems,
}

\DeclareAcronym{DT}{
  short = DT,
  long  = Digital Twin,
  short-plural-form = DTs,
  long-plural-form  = Digital Twins,
}

\DeclareAcronym{FM}{
  short = FM,
  long  = Foundation Model,
  short-plural-form = FMs,
  long-plural-form  = Foundation Models,
}

\DeclareAcronym{HMI}{
  short = HMI,
  long  = Human--Machine Interaction,
}

\DeclareAcronym{IoT}{
  short = IoT,
  long  = Internet of Things,
}

\DeclareAcronym{LLM}{
  short = LLM,
  long  = Large Language Model,
  short-plural-form = LLMs,
  long-plural-form  = Large Language Models,
}

\DeclareAcronym{MCP}{
  short = MCP,
  long  = Model Context Protocol,
}

\DeclareAcronym{MAS}{
  short = MAS,
  long  = Multi-Agent System,
  short-plural-form = MASs,
  long-plural-form  = Multi-Agent Systems,
}

\DeclareAcronym{PRISMA}{
  short = PRISMA,
  long  = Preferred Reporting Items for Systematic Reviews and Meta-Analyses,
}

\DeclareAcronym{RAMI}{
  short = {RAMI~4.0},
  long  = Reference Architecture Model Industry~4.0,
}

\DeclareAcronym{RAG}{
  short = RAG,
  long  = Retrieval-Augmented Generation,
}

\DeclareAcronym{TRL}{
  short = TRL,
  long  = Technology Readiness Level,
  short-plural-form = TRLs,
  long-plural-form  = Technology Readiness Levels,
}

\DeclareAcronym{MLLM}{
  short = MLLM,
  long  = Multimodal Large Language Model,
  short-plural-form = MLLMs,
  long-plural-form  = Multimodal Large Language Models,
}

\acsetup{list/display=used}

\usepackage{tikz}
\usetikzlibrary{shapes.geometric,arrows.meta}

\usepackage[most]{tcolorbox}
\newtcolorbox{definitionbox}[1][]{
  colback=white,
  colframe=gray!50,
  boxrule=0.4pt,
  arc=2pt,
  left=6pt,
  right=6pt,
  top=4pt,
  bottom=4pt,
  title={#1},
  fonttitle=\bfseries,
  coltitle=black
}

\newcommand{\defwool}[1]{\sethlcolor{teal!22}\hl{#1}}
\newcommand{\defvdi}[1]{\sethlcolor{orange!12}\hl{#1}}
\newcommand{\deffm}[1]{\sethlcolor{violet!12}\hl{#1}}
\newcommand{\defcap}[1]{\sethlcolor{yellow!12}\hl{#1}}

\def\tsc#1{\csdef{#1}{\textsc{\lowercase{#1}}\xspace}}
\tsc{WGM}
\tsc{QE}

\makeatletter
\AtBeginDocument{%
  \providecommand{\blocation}[1]{#1}%
  \let\blocation@orig\blocation
  \renewcommand{\blocation}[1]{%
    \def\@bltmp{#1}\def\@blqqq{???}%
    \ifx\@bltmp\@blqqq\else\blocation@orig{#1}\fi%
  }%
}
\makeatother

\begin{document}

\title[FM-Based Agents in Industrial Automation]{Foundation-Model-Based Agents in Industrial Automation: Purposes, Capabilities, and Open Challenges}

\author*[1]{\fnm{Vincent} \sur{Henkel}}\email{vincent.henkel@hsu.hamburg}
\equalcont{These authors contributed equally to this work.}
\author[1]{\fnm{Felix} \sur{Gehlhoff}}\email{felix.gehlhoff@hsu.hamburg}
\equalcont{These authors contributed equally to this work.}
\author[2]{\fnm{David} \sur{Kube}}\email{david.kube@siemens.com}
\author[3]{\fnm{Asaad} \sur{Almutareb}}\email{asaad.almutareb@artiquare.com}
\author[4]{\fnm{Luis} \sur{Cruz}}\email{luicruz@uan.edu.co}
\author[5]{\fnm{Bernd} \sur{Hellingrath}}\email{bernd.hellingrath@ercis.uni-muenster.de}
\author[6]{\fnm{Philip} \sur{Koch}}\email{philip.koch@ifam.fraunhofer.de}
\author[7]{\fnm{Christoph} \sur{Legat}}\email{christoph.legat@tha.de}
\author[8]{\fnm{Florian} \sur{Mohr}}\email{f.mohr@umwelt-campus.de}
\author[9]{\fnm{Michael} \sur{Oberle}}\email{michael.oberle@ipa.fraunhofer.de}
\author[10]{\fnm{Felix} \sur{Ocker}}\email{felix.ocker@honda-ri.de}
\author[11]{\fnm{Thorsten} \sur{Schoeler}}\email{thorsten.schoeler@tha.de}
\author[12]{\fnm{Mario} \sur{Thron}}\email{mario.thron@ifak.eu}
\author[6]{\fnm{Nico Andre} \sur{Töpfer}}\email{nico.andre.toepfer@ifam.fraunhofer.de}
\author[13]{\fnm{Lucas} \sur{Vogt}}\email{lucas.vogt@tu-dresden.de}
\author[14]{\fnm{Yuchen} \sur{Xia}}\email{yuchen.xia@ias.uni-stuttgart.de}

\affil*[1]{%
  \orgdiv{Institute of Automation Technology},
  \orgname{Helmut Schmidt University / University of the Federal Armed Forces Hamburg},
  \orgaddress{\city{Hamburg}, \country{Germany}}%
}


\affil[2]{%
  \orgname{Siemens AG},
  \orgaddress{\city{Nuremberg}, \country{Germany}};
  \orgdiv{Institute for Technologies and Management of Digital Transformation},
  \orgname{Bergische Universität Wuppertal},
  \orgaddress{\country{Germany}}%
}

\affil[3]{%
  \orgname{Artiquare GmbH},
  \orgaddress{\city{Ingolstadt}, \country{Germany}}%
}

\affil[4]{%
  \orgdiv{Facultad de Ingeniería Mecánica, Electrónica y Biomédica (FIMEB)},
  \orgname{Universidad Antonio Nariño},
  \orgaddress{\city{Bogotá}, \country{Colombia}}%
}

\affil[5]{%
  \orgdiv{Chair of Information Systems and Supply Chain Management},
  \orgname{University of Münster},
  \orgaddress{\city{Münster}, \country{Germany}}%
}

\affil[6]{%
  \orgname{Fraunhofer Institute for Manufacturing Technology and Advanced Materials IFAM},
  \orgaddress{\city{Stade}, \country{Germany}}%
}

\affil[7]{%
  \orgdiv{Research Group on Cognitive Autonomy \& Predictive Intelligence, Faculty of Electrical Engineering},
  \orgname{Technical University of Applied Sciences Augsburg},
  \orgaddress{\city{Augsburg}, \country{Germany}}%
}

\affil[8]{%
  \orgdiv{Birkenfeld Institutes of Technology},
  \orgname{Trier University of Applied Sciences},
  \orgaddress{\city{Birkenfeld}, \country{Germany}}%
}

\affil[9]{%
  \orgname{Fraunhofer Institute for Manufacturing Engineering and Automation IPA},
  \orgaddress{\city{Stuttgart}, \country{Germany}}%
}

\affil[10]{%
  \orgname{Honda Research Institute Europe},
  \orgaddress{\city{Offenbach am Main}, \country{Germany}}%
}

\affil[11]{%
  \orgdiv{Faculty of Computer Science},
  \orgname{Augsburg Technical University of Applied Sciences},
  \orgaddress{\city{Augsburg}, \country{Germany}}%
}

\affil[12]{%
  \orgname{Institute for Automation and Communication (ifak)},
  \orgaddress{\city{Magdeburg}, \country{Germany}}%
}

\affil[13]{%
  \orgdiv{Process-to-Order Group},
  \orgname{TUD Dresden University of Technology},
  \orgaddress{\city{Dresden}, \country{Germany}}%
}

\affil[14]{%
  \orgdiv{Institute for Industrial Automation and Software Engineering},
  \orgname{University of Stuttgart},
  \orgaddress{\city{Stuttgart}, \country{Germany}}%
}

\abstract{Foundation models, particularly large language models, are increasingly integrated into agent architectures for industrial tasks such as decision support, process monitoring, and engineering automation. Yet evidence on their purposes, capabilities, and limitations remains fragmented across domains. This work examines how mature foundation-model-based agent systems are in industrial contexts, how their functional profile differs from conventional agent systems, and which limitations persist. A systematic literature survey following the PRISMA~2020 guideline is presented, screening 2\,341 publications and synthesising a corpus of $88$ publications through a structured coding scheme. The results show that reported systems are predominantly at prototype and early validation stages (75.0\,\% at TRL~4--6), with deployment-oriented evidence remaining rare (9.1\,\%). Operational goals are most frequently positioned in user assistance, monitoring, and process optimisation, while conventional production-control purposes such as planning and scheduling are less prominent. Compared with an established baseline for industrial agent systems, the capability profile reveals substantial gains in human interaction (+37\,\%) and dealing with uncertainty (+35\,\%), but a pronounced deficit in negotiation ($-$39\,\%). The most widely reported limitations concern lack of generalization, hallucination and output instability, data scarcity, and inference latency. A working definition of \emph{foundation-model-based industrial agents} is also proposed, bridging conventional agent theory, automation-engineering standards, and the foundation-model paradigm.}

\keywords{foundation models, large language models, multi-agent systems, industrial automation, systematic literature review}

\maketitle

\printacronyms[heading=section*,name=Abbreviations]

\section{Introduction}\label{sec:introduction}
Software agents and \acp{MAS} have been studied for decades as a design paradigm for distributed decision-making in industrial domains such as production control, logistics, and process engineering~\cite{wooldridge2002,reinpold2025agentsDTcomparison}. In conventional settings, i.e. agent systems that predate the integration of \acp{FM}, agents are typically rule-based or optimisation-driven entities that pursue locally specified objectives under predefined interaction protocols such as the Contract Net Protocol~\cite{smith1980contractnet}. With the emergence of \acp{FM}, and \acp{LLM} in particular, a new class of agent systems has gained momentum~\cite{xi2025riseLLMagents}. Unlike their conventional counterparts, \ac{FM}-based agents can interpret unstructured and noisy information such as natural-language instructions, maintenance logs, and visual or sensor-stream inputs, interact with human operators through conversational interfaces, and orchestrate heterogeneous tool chains by generating executable code or application programming interface calls through flexible reasoning~\cite{wang2024surveyLLMagents}.
According to~\citet{ren2025aiagents}, \acp{FM}-based industrial agents comprise different levels of technological capabilities: \ac{LLM}-agents primarily extend language-centric reasoning and tool use, \ac{MLLM}-agents add multimodal perception across textual, visual, and sensor data, whereas Agentic AI refers to a further step toward self-directed, goal-driven autonomy in dynamic environments.
The shift from conventional \ac{MAS} to \ac{FM}-based agent systems is not only a shift in realised capabilities but also in the underlying coordination logic. \citet{abouali2026agenticAI}~point out the constrast between symbolic coordination through explicit protocols such as the Contract Net Protocol or blackboard systems with neural coordination based on structured conversation, role-based workflows, and prompt-driven orchestration.

Thus, even though there have been attempts to address challenges such as interaction with human operators, tool orchestration, and general optimisation applications in the pre-\ac{FM} era, these new technologies make such systems much more accessible, easier to develop and maintain, as well as considerably more capable and adaptive.

These developments have led to rapid adoption across a broad range of industrial applications, from engineering design automation and shop-floor control to energy-system operation and information-technology infrastructure management~\cite{Xia2026,deng2024maintenancedecisionsupport,gamage2024multiagentragenergy}. At the architectural level, \ac{LLM}-based agents are commonly integrated as central cognitive components that interpret context, generate plans or recommendations, and invoke external tools, often augmented by retrieval mechanisms to ground decisions in domain-specific knowledge~\cite{wang2024surveyLLMagents,gamage2024multiagentragenergy}. Recent surveys on \ac{LLM}-based agents in general-purpose settings have mapped these architectural patterns, reasoning strategies, and tool-use mechanisms~\cite{wang2024surveyLLMagents,xi2025riseLLMagents,guo2024llmMultiAgents}. However, the degree to which these general findings transfer to industrial contexts, where safety, determinism, and integration with legacy systems impose additional constraints, remains an open question~\cite{perezcerrolaza2024safetycriticalAI}.

Despite this growing adoption, there is no established working definition that bridges conventional agent concepts and standards, such as autonomous action in an environment~\cite{wooldridge2002} or encapsulated entities with control objectives~\cite{vdivde2653}, with the \ac{FM} paradigm, resulting in possible confusion with approaches that merely use \acp{LLM} for text generation or conversational interaction. Recent surveys acknowledge this gap: \citet{jin2024llmAgentsSE} note that among researchers there is no ``clear distinction between LLMs and LLM-based agents'' and that ``unified standard and benchmarking'' remain in an early stage, while \citet{zhou2024taxonomyFMagents} call for a ``unified taxonomy'' to address the currently ``fragmented approach'' to classifying \ac{FM}-based agent architectures. A related problem refers to ``conceptual retrofitting'', i.e., the tendency to describe modern LLM-based systems using classical agent concepts such as Belief-Desire-Intention (BDI) or perceive--plan--act--reflect loops, despite substantial differences in their operational mechanisms~\cite{abouali2026agenticAI}. This lack of conceptual consolidation is also evident in recent manufacturing-focused literature, which explicitly notes that the definitions, capability boundaries, and interconnections of LLM-agents, \ac{MLLM}-agents, and Agentic AI remain insufficiently clarified~\cite{ren2025aiagents}. Compact operational characterisations have been proposed, e.g., defining agentic \acp{LLM} as systems that ``reason, act, and interact''~\cite{plaat2025agenticLLMs}, yet no consolidated definition exists that integrates these perspectives with automation-engineering standards. Without such a definition, a systematic literature review lacks a reproducible inclusion criterion.

At the same time, the level of maturity of \ac{FM}-based agents is unclear, since evidence on such systems is fragmented across application domains, levels of technological maturity, and evaluation practices. Domain-specific surveys confirm that the integration of \acp{LLM} in manufacturing is ``still in its initial stages''~\cite{zhang2024llmManufacturingSurvey}, and a recent meta-survey on agent evaluation describes the field as a ``complex and underdeveloped area'' and aims to bring ``clarity to the fragmented landscape of agent evaluation''~\cite{mohammadi2025evalBenchmarkAgents}. Individual publications report prototypes in manufacturing~\cite{Xia2026,lim2024llmmasmanufacturing}, logistics~\cite{13}, energy systems~\cite{gamage2024multiagentragenergy}, or engineering design~\cite{deng2024maintenancedecisionsupport}, but no consolidated overview maps the capabilities and maturity of these approaches to a common framework. Another study points out that existing industrial \ac{FM}-based agent approaches appear to cluster around assistive and task-oriented roles, whereas stronger forms of self-directed goal formulation and system-level orchestration are still largely discussed as an emerging Agentic AI vision rather than as established industrial practice~\cite{ren2025aiagents}. As a consequence, it remains difficult to assess the overall technological maturity of the field, to identify recurring limitations across domains, and to establish a comparable evaluation basis, leaving practitioners in doubt about the capabilities and merits of adoption of \ac{FM}-based agent systems in industrial contexts.

Furthermore, it is unclear how \ac{FM} integration changes the functional profile of industrial agents compared to classical \ac{MAS}. Earlier systematic work on software agents in industrial production~\cite{reinpold2025agentsDTcomparison} provides a baseline of purposes (e.g., planning, scheduling, control) and capabilities (e.g., negotiation, coordination, reactivity). Initial evidence suggests a shift: \citet{greis2025masLLMdigitalTwin} explicitly ``contrast the capabilities of classic autonomous software agents and LLM software agents'' in a digital-twin-enabled manufacturing context, and \citet{zhao2025llmMultiAgentManufacturing} observe that conventional agent negotiation relies on ``pre-defined and fixed heuristic rules'' that are ill-suited to dynamic disturbances, motivating a multimodal \ac{LLM}-based alternative. This suggests that the two paradigms may be complementary rather than \ac{FM}-based agents being the successor of conventional approaches. However, this has not been systematically examined across domains.

Against this background, the guiding research questions (RQs) of this work are:
\begin{itemize}
    \item \textbf{RQ1:} How can \ac{FM}-based industrial agents be defined, and what is the current maturity of such systems in industrial and industrially relevant research, including their technology readiness, application domains, and use cases?
    \item \textbf{RQ2:} Which system purposes and capabilities do \ac{FM}-based agents exhibit, and how does their functional profile differ from conventional industrial agent systems?
    \item \textbf{RQ3:} Which limitations, challenges, and future work directions are most frequently reported for \ac{FM}-based agent systems in industrial contexts?
\end{itemize}

To address these questions, this work follows a \ac{PRISMA}-style systematic review (Section~\ref{sec:method}). For each included publication, technology readiness, application domain, system purposes, capabilities, reported limitations, and future work directions are assessed and consolidated.

The main contributions of this work are threefold. First, it proposes a working definition of \ac{FM}-based technical agents that bridges conventional agent theory, automation-engineering standards, and the \ac{FM} paradigm, and evaluates whether this definition adequately captures the systems reported in the corpus (Section~\ref{subsec:discussion-definition}). Second, building on this definition, it provides a cross-domain synthesis of technological maturity, system purposes, and capability profiles of \ac{FM}-based agents, including a descriptive comparison with an established baseline for industrial agent systems. Third, it consolidates recurring limitations and future work directions into structured themes that can inform the design, evaluation, and deployment of \ac{FM}-based agent systems in industrial contexts.

\section{Related work}\label{sec:related-work}
This section relates the present work to three streams of prior research: industrial software agents and \acp{MAS} (Section~\ref{subsec:rw-industrial-agents}), \ac{FM}-based agents in industrial contexts (Section~\ref{subsec:rw-fm-agents}), and existing taxonomies for purposes and capabilities (Section~\ref{subsec:rw-taxonomies}). Section~\ref{subsec:rw-baseline} introduces the baseline used for comparative analysis.

\subsection{Industrial software agents and multi-agent systems}
\label{subsec:rw-industrial-agents}
Industrial software agents and \acp{MAS} have long been studied as a design paradigm for distributed decision-making and control in production and related industrial domains~\cite{XiLi17}. In conventional settings, agent-based approaches are typically motivated by the need to decompose complex objectives into locally manageable sub-problems (e.g., planning and scheduling, dispatching decisions, shop-floor control, and diagnosis and fault management) and to coordinate these decisions across heterogeneous resources under operational constraints~\cite{HSW24,gehlhoff2023dissertation}. At the same time, industrial applications impose strong non-functional requirements, including determinism, safety, and real-time suitability, which shape how autonomy and interaction mechanisms can be realized in practice~\cite{massouh2024safeReconfigurableMAS,cruzsalazar2022cppsAgentArchitecture}.

From a conceptual perspective, \citet{russellnorvig2020} provide a broad characterization of an agent as ``anything that can be viewed as perceiving its environment through sensors and acting upon that environment through actuators'', a definition that is deliberately paradigm-agnostic and accommodates both symbolic and subsymbolic realizations. Within the more specific \ac{MAS} tradition, \citet{wooldridge2002} emphasizes autonomous action, reactivity, and coordination among multiple entities, including interaction patterns such as negotiation where applicable. In industrial survey work, these conventional expectations are operationalized as concrete purposes and capabilities that can be evidenced from application reports in production contexts~\cite{reinpold2025agentsDTcomparison}. Traditional applications include, for example, heuristics-based decentralised process planning and scheduling as well as controlling resources and diagnosis tasks on the shop floor~\cite{gehlhoff2023dissertation}.

\subsection{Foundation-model-based agents in industrial contexts}
\label{subsec:rw-fm-agents}
Recent \ac{FM}-based industrial agent systems build on these foundations but often repurpose them towards assistive and decision-support-centric roles (e.g., maintenance decision support or manufacturing assistance) rather than towards fully decentralised negotiation-heavy control \cite{deng2024maintenancedecisionsupport,lim2024llmmasmanufacturing,Xia2026}. \acp{FM}, and in particular \acp{LLM}, provide generic language understanding and generation capabilities that can be adapted to various downstream tasks~\cite{bommasani2021opportunities}. In industrial and industrially relevant agent systems, \acp{LLM} are commonly integrated as central cognitive components that (i) interpret human instructions and contextual information, (ii) generate plans or recommendations, and (iii) orchestrate tool-mediated actions by interacting with external software components (e.g., databases, simulation models, code generators, or domain services)~\cite{chen2025llmDigitalTwinIndustry5}. A recurring architectural pattern is to complement \acp{LLM} with explicit grounding mechanisms such as \ac{RAG} to improve factuality and to connect agent decisions to domain-specific knowledge sources \cite{gamage2024multiagentragenergy}. \citet{ren2025aiagents}~also emphasize that industrial tasks also depend on the joint interpretation of, e.g., enterprise data, maintenance records, sensor streams, and machine-vision inputs, which is why multimodal \ac{LLM}-based agents are increasingly discussed as a relevant architectural extension that supports context-aware perception, diagnosis, and decision support.

Prior work has noted limitations related to robustness, reliability, and external validity, particularly when \acp{LLM} are used for decision-making under incomplete information or when generated outputs must be executable in technical systems. Practical constraints such as latency, cost, and integration effort are also discussed \cite{deng2024maintenancedecisionsupport,jee2025overlayissueclassification,lim2024llmmasmanufacturing,Xia2026}.

\subsection{Taxonomies for purposes, properties, and capabilities}
\label{subsec:rw-taxonomies}
Systematic comparison across agent systems requires a representation that separates \emph{what} a system is intended to achieve from \emph{how} it achieves it. 
 \citet{mueller2021industrialautonomoussystems}~characterise industrial autonomous systems through four higher-level dimensions---\emph{systematic process execution}, \emph{adaptability}, \emph{self-governance}, and \emph{self-containedness}---and relate these to lower-level abilities such as learning ability, decision-making capacity, cooperability, reactivity, and self-explanation. Since these relations are many-to-many rather than one-to-one, their framework suggests that higher-level system qualities should be analysed separately from the concrete functional means by which they are realised. 
 \citet{kaber2017conceptual} provides an additional conceptual basis for separating higher-level system qualities from concrete functionality by distinguishing \emph{automation} from \emph{autonomy}. Rather than treating autonomy as a higher level of automation, he defines it through three conditions---\emph{viability}, \emph{independence}, and \emph{self-governance}---and further operationalises the distinction in terms of the demands a system places on its environment, human collaborators, and task protocols. 
In this work, the analysis follows the purpose--property--capability framing employed by \citet{reinpold2025agentsDTcomparison}, which builds on the work by~\citet{mueller2021industrialautonomoussystems}. In this framing, \emph{system purposes} describe operational goal categories (e.g., planning, scheduling, control, monitoring, user assistance), \emph{capabilities} describe concrete functional skills evidenced by the system (e.g., interaction, communication, coordination, reasoning), and \emph{properties} aggregate capability evidence into higher-level dimensions that support corpus-level comparison (see Section~\ref{subsec:working-definition} for the working definition used in this review). This distinction is particularly useful for \ac{FM}-based agent systems because \acp{LLM} can simultaneously shift the operational emphasis towards assistive purposes and affect the evidencing of capabilities such as human interaction and uncertainty handling.

\subsection{Baseline for comparative analysis}
\label{subsec:rw-baseline}
\citet{reinpold2025agentsDTcomparison} systematically compare industrial software agents and \acp{DT} in production contexts through a \ac{PRISMA}-aligned literature review covering 145 publications. Their work provides the purpose and capability taxonomy adopted in this review (see Section~\ref{subsec:rw-taxonomies}) together with composite scores that summarize capability profiles at property level, thereby establishing a quantitative reference point for corpus-level comparison. Given the different temporal scopes of the two reviews, with \citet{reinpold2025agentsDTcomparison} largely predating the broad adoption of \acp{FM} in industrial agent research, substantial overlap between the two corpora is unlikely, and the two studies can be viewed as largely complementary samples of conventional and \ac{FM}-based agent research, respectively.

In this work, \citet{reinpold2025agentsDTcomparison} is used as a baseline for descriptive comparison, enabling the computation of comparable coverage profiles and differences in percentage points. Differences in corpus composition and inclusion criteria between the two studies should be considered when interpreting the comparison (see Section~\ref{subsec:discussion-validity}).
\section{Method}\label{sec:method}

This work follows the \ac{PRISMA}~2020 guideline for systematic literature reviews \cite{pagePRISMA2020Statement2021} and adapts it to the analysis of \ac{FM}-based agent systems in industrial and industrially oriented research. The procedure comprises the standard \ac{PRISMA} stages of (1) identification of potentially relevant records via a structured multi-database query and (2) relevance filtering according to predefined eligibility criteria, followed by (3) a consolidation step in which the resulting corpus is transformed into a consistent, machine-readable representation for quantitative analysis.

\subsection{Working definition}
\label{subsec:working-definition}

Agents have a long-standing history within academia~\cite{wooldridge2002}. While these classical foundations provide the essential basis for this review, the modern literature often uses the term ``agent'' interchangeably with related concepts such as agentic workflows or simply ``\acp{LLM}''. To maintain a clear scope, this work distinguishes its focus from these adjacent terms and provides a working definition that is (i) grounded in established agent concepts, (ii) compatible with automation-engineering notions of technical agents, and (iii) explicit about what it means for a system to be ``\ac{FM}-based''.

In the agent literature, \citet{wooldridge2002} defines an agent as \emph{``}\defwool{a computer system that is situated in some environment, and that is capable of autonomous action in this environment in order to meet its design objectives}\emph{''}. In the context of automation engineering, the VDI/VDE~2653 guideline defines a technical agent as \emph{``}\defvdi{an encapsulated hardware or software entity with specified objectives regarding the control of a technical system or a part thereof}\emph{''}~\cite{vdivde2653}. Foundation models, in turn, are characterized as models \emph{``}\deffm{trained on broad data (generally using self-supervision at scale) that can be adapted (e.g., fine-tuned) to a wide range of downstream tasks}\emph{''}~\cite{bommasani2021opportunities}. According to~\citet{ren2025aiagents}, systems with high degrees of autonomy that use these FMs exhibit capabilities such as \defcap{semantic retrieval and context-awareness, adaptive reasoning, and autonomous decision-making, which includes the realisation of selected actions}, i.e. influencing the actual process.

Combining these streams and concepts, the following definition is proposed that guides the focus of this work:

\begin{definitionbox}[Working definition used in this review]
\textbf{A foundation-model-based industrial agent} is
\defvdi{an encapsulated hardware or software entity acting in the context of an industrial system}
\defwool{that is capable of autonomous action in order to meet its specified design objectives},
and \deffm{that uses a foundation model as a central component} \defcap{for context interpretation, decision-making, as well as action selection and execution}.
\end{definitionbox}

Throughout the review, each category is assessed based on paper-level evidence: if the publication does not provide sufficient detail, the corresponding category is treated as not evidenced.

\subsection{Data sources and search strategy}
\label{subsec:data-sources}
Records were retrieved from four bibliographic and preprint sources: \textsc{Scopus}, \textsc{Semantic Scholar}, \textsc{arXiv}, and \textsc{OpenAlex}. The search covered publications from 2020 onwards and targeted \acp{LLM} and related \acp{FM} in agent or multi-agent settings within industrial and industrially relevant domains, including manufacturing, logistics, energy systems, and the engineering life cycle (product development and process engineering), as well as cross-domain functionalities such as maintenance, quality management, and \ac{HMI}. The core query combined (i) terms for \acp{LLM} and \acp{FM}, (ii) terms for agents and \acp{MAS}, and (iii) terms for industrial contexts. Across all sources, 3025 valid results and 2341 unique records were obtained.

The composite search query is shown in Table~\ref{tab:search-query}. The search and export were executed on September~8,~2025.

\begin{table*}[!htbp]
\centering
\caption{Search query used for multi-database retrieval. The query is composed of three conceptual blocks connected by Boolean \textsc{and}.}
\label{tab:search-query}
\small
\begin{tabular}{@{}p{0.10\linewidth}p{0.86\linewidth}@{}}
\toprule
\textbf{Block} & \textbf{Terms} \\
\midrule
\acs{FM}/\acs{LLM} & \texttt{"LLM" OR "large language model*" OR "foundation model*" OR "generative AI" OR "agentic AI" OR AI OR "Diffusion Model" OR "Vision Language Action Model" OR VLA OR "vision-language-action model" OR "World Model"} \\[4pt]
Agent & \texttt{agent OR "intelligent agent" OR "multi-agent system" OR MAS} \\[4pt]
Industrial context & \texttt{industr* OR manufactur* OR production OR factory OR shopfloor OR automat* OR logistic* OR mainten* OR quality OR "process engineering" OR "process industry" OR energy OR "energy system*" OR develop* OR "human-machine interaction" OR HMI OR "cyber-physical system*" OR "Simulation Model"} \\
\bottomrule
\end{tabular}
\end{table*}

\subsection{Screening and eligibility}
\label{subsec:screening}
Eligibility is defined at the level of individual publications. A record is included if it meets all of the following criteria:
\begin{enumerate}
    \item The publication describes a concrete application, implementation, or case study (not a purely conceptual contribution, high-level vision paper, survey, or review without practical realisation or empirical results).
    \item The system uses one or more \acp{LLM}, \acp{MLLM}, or related \acp{FM} for control or decision-making within an agent or agent-based architecture.
    \item The application context is industrial or industrially relevant, covering the above-mentioned domains, engineering life cycle phases, or cross-domain functionalities.
\end{enumerate}

Given the size of the initial corpus, the title and abstract screening was supported by an \ac{LLM}, which assigned each record a relevance score against the eligibility criteria stated above. The inclusion threshold was calibrated on manually annotated samples, and a subset of below-threshold records was reviewed manually to control for false negatives. Records above the threshold were forwarded to full-text assessment, where final eligibility was determined by the authors.

After title and abstract screening, a full-text assessment determines whether the use of an agent or \ac{MAS} qualifies as a \emph{foundation-model-based technical agent} according to the working definition in Section~\ref{subsec:working-definition}. At this stage, publications that constitute surveys or reviews without an original application contribution were excluded. The remaining full texts were assessed against the working definition, verifying that the system exhibits autonomous behaviour,  beyond pure text generation and that a \ac{FM} is involved in the decision loop (e.g., interpreting context, planning, selecting actions, or orchestrating tool calls).

\begin{figure}[!htbp]
    \centering
    \resizebox{\linewidth}{!}{%
        \begin{tikzpicture}[
    node distance=0.9cm and 0.6cm,
    box/.style={rectangle, draw, fill=blue!10, text width=2.8cm,
                text centered, minimum height=1.1cm,
                font=\scriptsize},
    reason/.style={rectangle, draw, fill=blue!10, text width=3.8cm,
                   text badly centered, minimum height=1.1cm,
                   font=\scriptsize},
    arrow/.style={-{Stealth[length=5pt]}, thick, blue!70},
    label/.style={font=\scriptsize, text=blue!70, fill=white, inner sep=1pt}
]

\node[box] (identify)
    {\textbf{Identification}\\Records identified\\across all sources\\$n = 3{,}025$};

\node[box, below=of identify] (screen)
    {\textbf{Screening}\\Unique records after\\deduplication\\$n = 2{,}341$};

\node[box, below=of screen] (eligible)
    {\textbf{Eligibility}\\Records screened for\\relevance (title/abstract)};

\node[box, below=of eligible] (include)
    {\textbf{Included}\\Studies included in\\the final corpus\\$n = 88$};

\node[reason, right=of eligible] (reason1) {%
    \textbf{Excluded ($n = 2{,}253$)}\\[3pt]
    {\small\setlength{\leftmargini}{10pt}%
    \raggedright
    \begin{itemize}\setlength\itemsep{0pt}\setlength\parsep{0pt}
      \item No industrial or industrially relevant context
      \item Conceptual only, no implementation
      \item Review or survey without application
    \end{itemize}}};

\draw[arrow] (identify) -- (screen);
\draw[arrow] (screen)   -- (eligible);
\draw[arrow] (eligible) -- (include);

\draw[arrow] (eligible.east) -- (reason1.west);

\end{tikzpicture}
    }
    \caption{\acs{PRISMA}-style flow of records for the identification, screening, and inclusion of publications on \ac{FM}-based agent systems in industrial and industrially relevant contexts.}
    \label{fig:prisma-flow}
\end{figure}
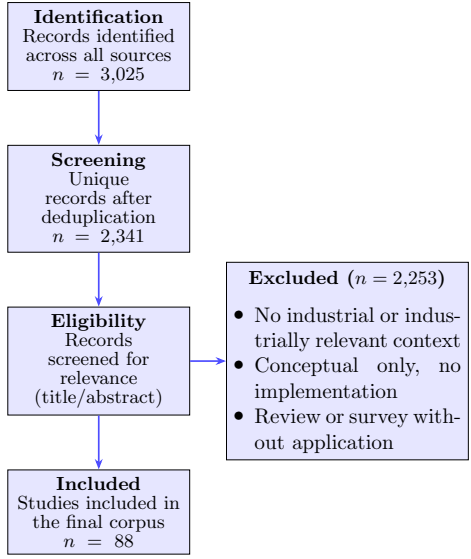
\FloatBarrier

The overall identification, screening, and inclusion process is summarized in the \acs{PRISMA} flow diagram in Figure~\ref{fig:prisma-flow}. The title and abstract screening have been supported by an AI tool chain, i.e. an LLM was used to assess the probability of a title matching the above-stated criteria. Afterwards, the threshold has been determined by looking at different samples, resulting in a corpus comprising $N=88$ included publications. By publication type, conference papers dominate ($49/88$, 55.7\%), followed by preprints ($20/88$, 22.7\%) and journal articles ($19/88$, 21.6\%). Nearly all publications date from 2024 ($38/88$) or 2025 ($42/88$), confirming that \ac{FM}-based agent systems in industrial contexts are a rapidly emerging research area.

\subsection{Taxonomies and operationalization}
\label{subsec:taxonomies-operationalization}
Following the purpose--property--capability framing introduced in Section~\ref{subsec:rw-taxonomies}, this work adopts the taxonomy of \citet{reinpold2025agentsDTcomparison} to enable a transparent and comparable characterization of agent systems across publications.

Based on this distinction, each included publication is mapped to a fixed taxonomy of operational purposes (planning, scheduling, dispatching, control, diagnosis and fault management, user assistance, monitoring, virtual commissioning, and process optimisation). In addition, reported system capabilities are classified according to the capability taxonomy and aggregated into higher-level property dimensions (sociability, autonomy, intelligence, and fidelity) to support corpus-level comparison. Classification is performed at paper level: each category is marked as \emph{supported} if explicit textual evidence is provided in the publication, and as \emph{not evidenced} otherwise. Because a single system can serve multiple operational goals and exhibit multiple capabilities, purpose and capability assignments are treated as multi-label indicators.

\subsection{Comparative analysis and aggregation}
\label{subsec:comparative-analysis}
Comparative analysis against the baseline survey~\cite{reinpold2025agentsDTcomparison} is performed in a descriptive manner. Using the shared taxonomy, relative occurrence shares are compared by reporting differences in percentage points. Higher-level property dimensions are computed as weighted averages of the underlying capability indicators, following the weighting scheme defined by \citet{reinpold2025agentsDTcomparison}. Reported limitations and future work directions are grouped into thematic categories.

\section{Results}\label{sec:results}
This section details the results of the corpus analysis ($N=88$), structured along the three research questions introduced in Section~\ref{sec:introduction}. Section~\ref{subsec:results-rq1} addresses RQ1 by establishing the descriptive statistics of the field, focusing on \acp{TRL} and application domains. Section~\ref{subsec:results-rq2} addresses RQ2 through a comparative analysis of system purposes and capabilities. Finally, Section~\ref{subsec:results-rq3} addresses RQ3 by consolidating the reported limitations and future work directions.

\subsection{Technological maturity and application domains}
\label{subsec:results-rq1}
Addressing RQ1, this section characterises the technological maturity of the corpus along the \ac{TRL} scale and its distribution across application domains.

\begin{figure}[!htbp]
    \centering
\includegraphics[width=\linewidth]{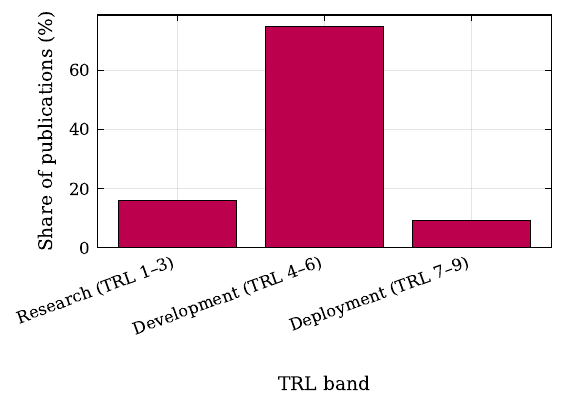}
    \caption{Aggregated technology readiness levels (\acs{TRL} bands) of \acs{FM}-based technical agents across the corpus.}
    \label{fig:trl-bands}
\end{figure}
Technology readiness is assessed using the nine-level \acs{TRL} scale~\cite{mankins1995trl}, aggregated into three bands: research (\acs{TRL}~1--3, basic principles to proof of concept), development (\acs{TRL}~4--6, laboratory validation to demonstration), and deployment (\acs{TRL}~7--9, system prototype to operational use).

As shown in Figure~\ref{fig:trl-bands}, research-stage contributions (\acs{TRL}~1--3) account for only $14/88$ (15.9\%). This low share should be read in light of the inclusion criteria (Section~\ref{subsec:screening}), which require a concrete application, implementation, or case study and therefore tend to filter out purely conceptual contributions. Beyond this, the relatively high share of \acs{TRL}~4+ systems suggests that laboratory-stage prototypes are commonly reached and feasible with \ac{FM}-based approaches. Accordingly, the majority of reported systems fall into the development band (\acs{TRL}~4--6; $66/88$, 75.0\%), indicating that most contributions present laboratory prototypes or early-stage validations. Deployment-oriented evidence (\acs{TRL}~7--9), however, remains rare ($8/88$, 9.1\%), showing that despite the ability to quickly develop lab-scale demonstrators, the step to deployment in operation is more difficult and has rarely been achieved.

To characterize the industrial landscape addressed by the corpus, each publication is assigned to one of four application domains. The first three domains are inspired by the application scenarios of \emph{Plattform Industrie~4.0}~\cite{plattformi40landkarte2018}; the fourth was added inductively to accommodate energy systems, IT operations, and transportation use cases that fall outside the original manufacturing-centric scope:

\begin{description}
    \item[Production \& Manufacturing Control] covers planning, control, and optimisation of production and manufacturing processes.
    \item[Logistics \& Supply Chain] covers coordination, decision support, and automation in logistics and supply chain operations.
    \item[Engineering \& Design Automation] covers technical development, simulation, design automation, and engineering workflows.
    \item[Industrial Operations \& Networks] covers energy systems, grid and network operations, IT and cloud operations, transportation, and cross-domain industrial decision support.
\end{description}

\begin{figure}[!htbp]
    \centering
    \includegraphics[width=\columnwidth]{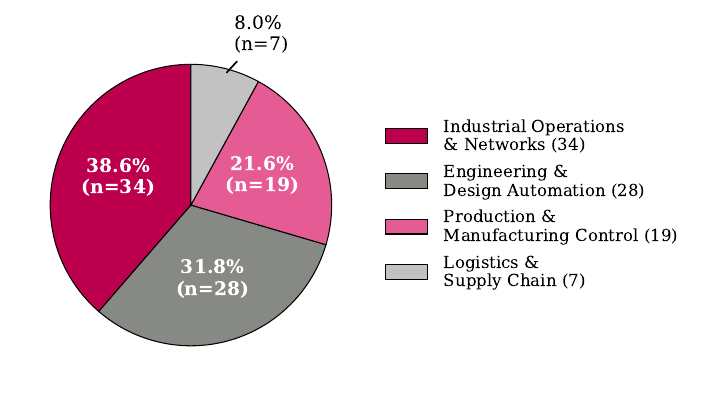}
    \caption{Distribution of application domains across the corpus ($N=88$).}
    \label{fig:domain-distribution}
\end{figure}

As shown in Figure~\ref{fig:domain-distribution}, the largest cluster is \emph{Industrial Operations \& Networks} ($34/88$, 38.6\%), covering power grid operation and voltage control~\cite{8,11,39,40,62,74}, energy management and charging infrastructure~\cite{64,69,91,96}, cloud and IT operations~\cite{7,19,71,77,88}, transportation and autonomous vehicles~\cite{14,60,80}, building management~\cite{1,92}, telecommunications~\cite{67,68}, and cross-domain decision support for industrial knowledge systems~\cite{30,32,33,42,46,65,66,72,75,82,93}. The second-largest cluster is \emph{Engineering \& Design Automation} ($28/88$, 31.8\%), including electronic design automation and analog circuit layout~\cite{37,45,59,63,84}, computational simulation and modelling~\cite{25,26,41,44,47}, product and mechanical design~\cite{28,50,57,83,86,95}, process automation and knowledge-assisted workflows~\cite{24,38,49,54,55,56,58,81,94}, and maintenance decision support and risk assessment~\cite{22,35,85}. \emph{Production \& Manufacturing Control} accounts for $19/88$ papers (21.6\%), spanning shop-floor control and multi-agent manufacturing systems~\cite{6,18,20,21,27,43,79}, robotic manipulation and human--robot collaboration~\cite{4,51,61}, semiconductor manufacturing~\cite{17,34}, process planning~\cite{9,23,53}, predictive maintenance~\cite{3}, mining~\cite{16}, agricultural data management~\cite{70}, and materials discovery and industrial applications~\cite{78,87}. Finally, \emph{Logistics \& Supply Chain} forms a smaller but distinct group ($7/88$, 8.0\%), addressing fleet dispatching~\cite{13}, warehouse management~\cite{15}, heterogeneous multi-agent coordination for delivery and rescue~\cite{5,12,31,48}, and procurement~\cite{36}.

\subsection{System purposes and capabilities}
\label{subsec:results-rq2}
Addressing RQ2, this section compares the purpose and capability profile of \ac{FM}-based agents with the baseline survey~\cite{reinpold2025agentsDTcomparison}. 

\begin{figure}[!htbp]
    \centering
    \includegraphics[width=1\columnwidth]{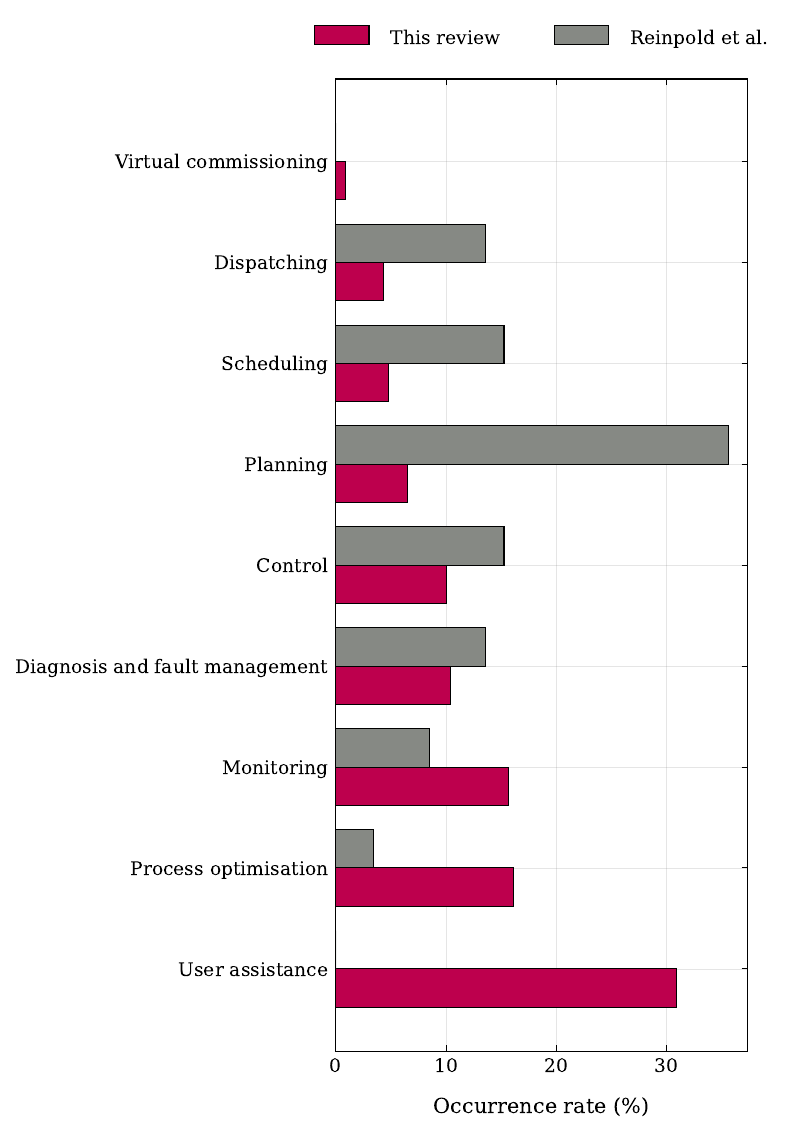}
    \caption{Relative occurrence of system purposes in the coded corpus compared to the reference taxonomy from \citet{reinpold2025agentsDTcomparison}.}
    \label{fig:system-purposes}
\end{figure}

Figure~\ref{fig:system-purposes} compares the purpose profile of \ac{FM}-based agents with the baseline survey~\cite{reinpold2025agentsDTcomparison}. Because each publication may serve multiple operational goals, purposes are coded as multi-label indicators; the corpus yields 230 purpose assignments in total. User assistance accounts for the largest share (71 assignments, 30.9\%), whereas it is absent in the baseline (+30.9\,pp). This indicates a notable shift in the orientation of agent-based applications. Classical agent surveys have not reported user assistance as a system purpose. The integration of \acp{FM} into industrial agents appears to have enabled this category, thereby enlarging the set of applications that can be addressed. Process optimisation (37 assignments, 16.1\%; baseline 3.4\%) and monitoring (36 assignments, 15.7\%; baseline 8.5\%) follow as the second and third most frequent purposes. Here as well, a considerable increase in the utilisation of agent-based systems is observed. Note that process optimisation here refers to the improvement of resource design and performance, not to the application of, e.g., optimisation heuristics to optimize a whole production process ~\cite{reinpold2025agentsDTcomparison}. The findings imply a convergence of the agent and DT paradigm as traditionally, these purposes have largely been fulfilled by DT applications~\cite{reinpold2025agentsDTcomparison}. It appears that \acp{FM} enable agents to process and leverage domain-specific information, thus facilitating their use for these knowledge-intensive applications.

Conversely, conventional production-control purposes are markedly less prominent than in the baseline. Planning drops from 35.6\% to 6.5\% (15 assignments), scheduling from 15.3\% to 4.8\% (11), and dispatching from 13.6\% to 4.3\% (10). Here, classical heuristics and optimisation-focused approaches have been more pronounced. A possible reason for this drop in the number of planning-oriented approaches with FM-based agents is the limited reliability of agents with FMs (Figure~\ref{fig:limitations}) due to hallucinations and instabilities.  Control accounts for 23 assignments (10.0\%; baseline 15.3\%); while the absolute number of control-related papers is comparable, its relative share decreases because the overall set of purpose assignments is larger due to the emergence of new purposes such as user assistance. Diagnosis and fault management (24 assignments, 10.4\%; baseline 13.6\%; $-3.1$\,pp) and virtual commissioning (2, 0.9\%; baseline 0.0\%; +0.9\,pp) are retained at near-baseline levels, indicating that knowledge-intensive and verification-oriented purposes are preserved in \ac{FM}-based systems, unlike the sharper decline in planning, scheduling, and dispatching.

Overall, the purpose profile indicates a repositioning of agent functionality from embedded, optimisation-centred production control towards assistive, monitoring-oriented roles.

\begin{figure*}[!htbp]
    \centering
\includegraphics[width=\linewidth]{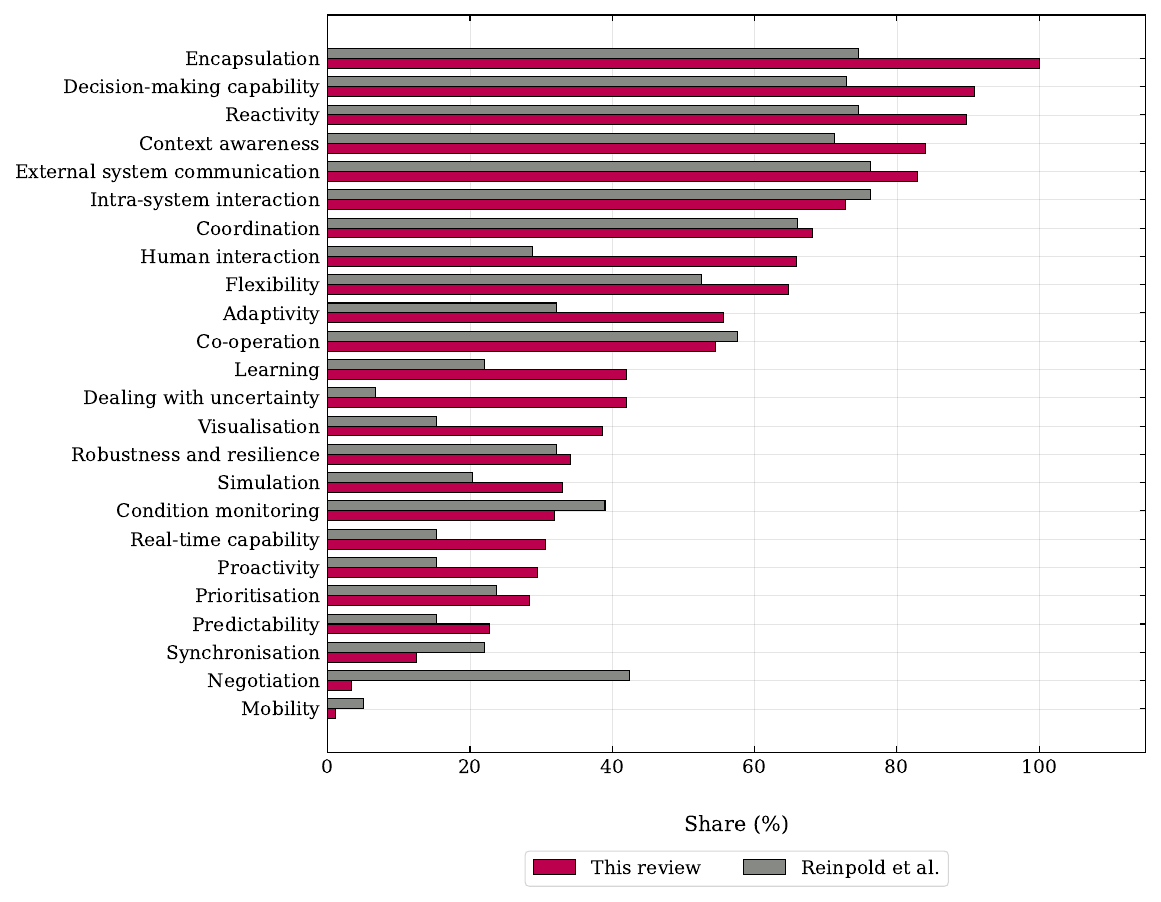}
    \caption{Capability coverage of \ac{FM}-based agent systems in this work compared with \citet{reinpold2025agentsDTcomparison}.}
    \label{fig:capabilities-comparison}
\end{figure*}

The capability profile in Figure~\ref{fig:capabilities-comparison} reveals both continuities and marked shifts relative to the baseline. The most frequently evidenced capabilities are encapsulation ($88/88$, 100\%; set to \emph{supported} by design, as the working definition requires encapsulated entities and the prompt-based interaction paradigm inherently hides internal states, see Section~\ref{subsec:discussion-definition}), decision-making ($80/88$, 90.9\%), reactivity ($79/88$, 89.8\%), context awareness ($74/88$, 84.1\%), external system communication ($73/88$, 83.0\%), and coordination ($60/88$, 68.2\%). Within this set of capabilities, the findings indicate a similar yet more pronounced capability pattern compared to classical industrial agents, where these capabilities are likewise the most frequently reported. Condition monitoring ($28/88$, 31.8\%), robustness and resilience ($30/88$, 34.1\%), prioritisation ($25/88$, 28.4\%), and mobility ($1/88$, 1.1\%) also show no major differences to the baseline. The similar score in robustness and resilience, i.e.\ the capability to react to disturbances and disruptive events, can be attributed to similar vulnerabilities within both architectural paradigms: in neither case do the majority of reported systems demonstrate graceful degradation or continued operation under partial component failure. It should be noted that the results reflect the empirical evidence reported in the publications, not the potential capabilities of the systems.

Marked differences emerge in capabilities that are directly enabled by \acp{FM}. Human interaction ($58/88$, 65.9\%; $+37.1$\,pp) marks the largest positive deviation, which is expected as FMs, especially LLMs, provide a much more accessible human–machine interface through interacting with the user in natural language. Dealing with uncertainty ($37/88$, 42.0\%; $+35.3$\,pp) shows the second-largest increase, indicating that \acp{FM} enable agents to reason and act under incomplete or ambiguous information, a capability that classical rule-based and optimisation-driven agents rarely exhibit. Adaptivity ($49/88$, 55.7\%; $+23.5$\,pp), learning ($37/88$, 42.0\%; $+20.0$\,pp), and proactivity ($26/88$, 29.5\%; $+14.3$\,pp) are likewise substantially more prevalent, implying that \acp{FM} as a data-driven approach, provide the technological means to enable these complex capabilities. Visualisation ($34/88$, 38.6\%; $+23.4$\,pp) and simulation ($29/88$, 33.0\%; $+12.6$\,pp), capabilities commonly associated with \acp{DT}, can be found more often in \ac{FM}-based approaches, where tool and model access is facilitated by standardised interfaces such as \ac{MCP}. Flexibility ($57/88$, 64.8\%) and co-operation ($48/88$, 54.5\%) remain at comparable levels to the baseline.

Interestingly, real-time capability ($27/88$, 30.7\%; $+15.4$\,pp) scores higher in \ac{FM}-based approaches. Here, real-time capability refers to the ability to react under tight temporal restrictions, not necessarily hard real-time constraints. The latter requires further investigation into the actual hard-real-time capabilities of these systems.

Notably, classical approaches score considerably higher in synchronisation ($11/88$, 12.5\%; $-9.5$\,pp) and especially negotiation ($3/88$, 3.4\%; $-39.0$\,pp). Synchronisation, i.e.\ the capability to react to changes in the system and update the agent's internal model, has been more prominent in classical approaches, likely because many optimisation algorithms require a synchronised model of the underlying system. The major difference in negotiation corresponds to the shift in application focus identified in the purpose analysis, showcasing the transition from negotiation-based optimisation heuristics to a more user-centric, assistive paradigm. The typically small number of agents observed in the corpus may additionally reduce the opportunities for negotiation by construction.

\begin{figure*}[!htbp]
    \centering
\includegraphics[width=0.8\linewidth]{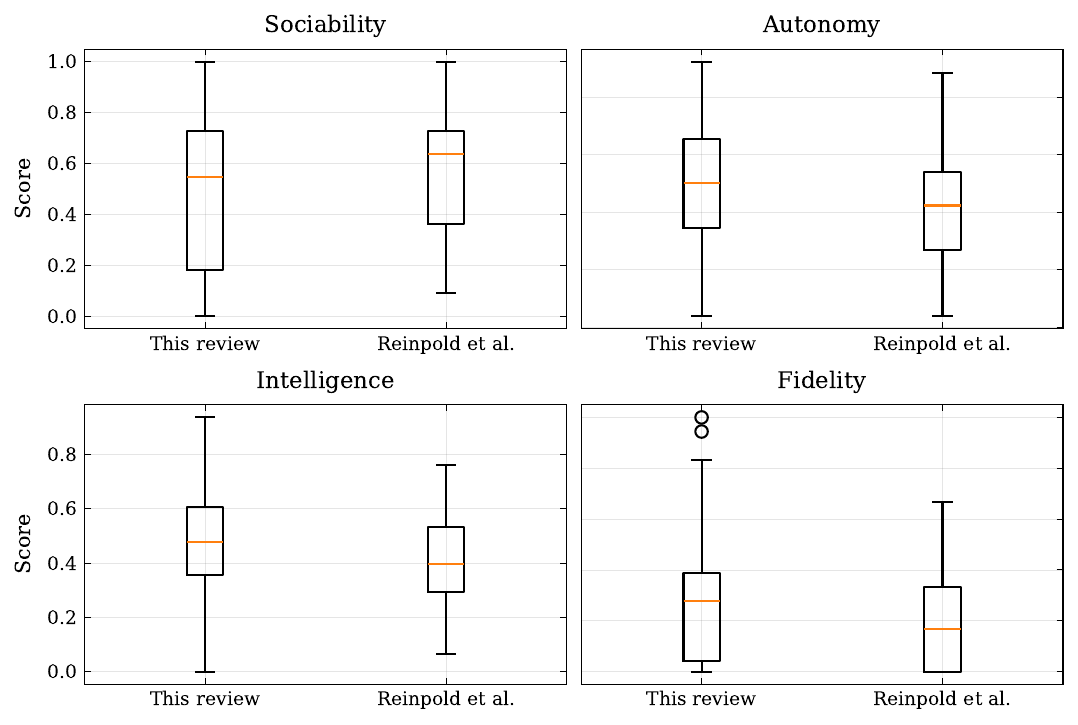}
    \caption{Composite scores for sociability, autonomy, intelligence, and fidelity comparing this work with \citet{reinpold2025agentsDTcomparison}.}
    \label{fig:composite-scores}
\end{figure*}

The composite score comparison in Figure~\ref{fig:composite-scores} aggregates these capability-level observations into higher-level property dimensions, following the weighting scheme defined in \citet{reinpold2025agentsDTcomparison}. Autonomy (median $0.50$ vs.\ $0.42$, $+0.08$), intelligence (median $0.48$ vs.\ $0.40$, $+0.08$), and fidelity (median $0.28$ vs.\ $0.17$, $+0.11$) are consistently higher in this work. Sociability, by contrast, shows a lower median ($0.55$ vs.\ $0.64$, $-0.09$), which is consistent with the reduced negotiation capability identified above.

\subsection{Limitations and future work}
\label{subsec:results-rq3}
Addressing RQ3, this section consolidates the most frequently reported limitations and future work directions across the corpus. 

\begin{figure}[!htbp]
    \centering
\includegraphics[width=\columnwidth]{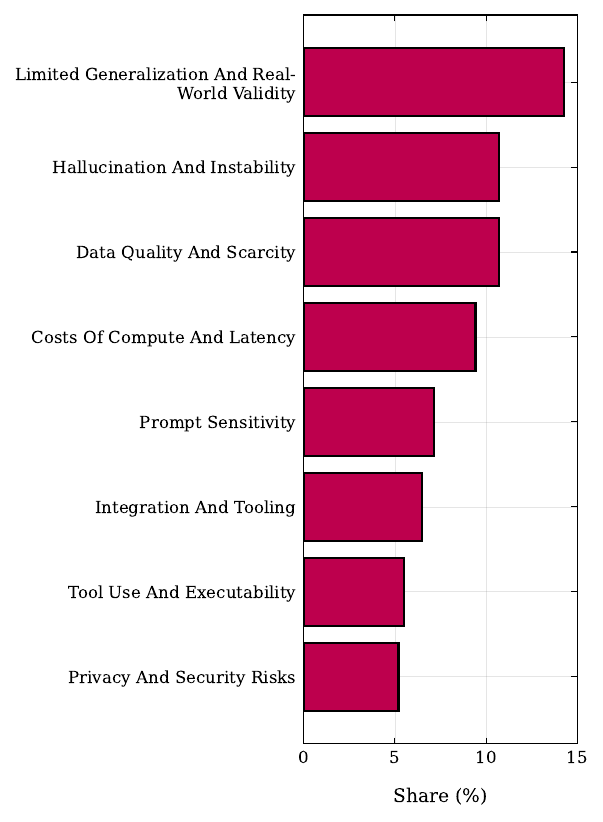}
    \caption{Top limitation themes for \ac{FM}-based agent systems in industrial and industrially relevant settings, measured by the share of papers in which each theme is reported.}
    \label{fig:limitations}
\end{figure}

Figure~\ref{fig:limitations} summarises the most frequently reported limitation themes across 308 total limitation mentions. Limited generalisation and real-world validity is the dominant concern (44 mentions, 14.3\%). This theme subsumes simulation-only evaluations, proof-of-concept scope, and restricted domain coverage: many systems are validated in controlled laboratory settings, on narrow benchmarks, or in single-site case studies, leaving open how well they transfer to different operational conditions or industrial environments. This finding is consistent with the \acs{TRL} distribution reported in Section~\ref{subsec:results-rq1}, where the majority of systems remain at development stage. Hallucination and instability (33, 10.7\%) ranks second and encompasses not only factual hallucinations but also non-deterministic behaviour across repeated runs, numeric and logical errors, and the generation of non-executable outputs. Data quality and scarcity (33, 10.7\%) captures domain-specific data deficits, reliance on synthetic or biased datasets, and challenges in obtaining accurate system parameters. Costs of compute and latency (29, 9.4\%) completes the top cluster and reflects inference latency, high computational cost, and limited scalability under concurrent workloads.

The remaining themes point to integration-layer and governance challenges. Prompt sensitivity (22, 7.1\%) highlights the brittleness of natural-language-mediated control: model behaviour is highly sensitive to prompt phrasing and ordering, decoding parameters, and formatting conventions, often requiring handcrafted or few-shot prompts and additional guardrails to produce valid and stable outputs. Integration and tooling (20, 6.5\%) addresses the complexity of retrofitting \acp{FM} into industrial toolchains, including vendor dependencies, format conversions between simulation, CAD, or \ac{DT} data and natural-language representations, and inconsistent interfaces across automation layers. Tool use and executability (17, 5.5\%) is a related but distinct concern: \ac{FM}-based agents frequently misuse tools or APIs, generate syntactically invalid code or structured outputs (e.g., JSON, SQL, domain-specific languages), and produce non-executable plans that require verification and retries. Privacy and security risks (16, 5.2\%) round out the profile and encompass data leakage through model queries, dependence on external API providers, and vulnerability to prompt injection and adversarial manipulation of \ac{FM}-based decision loops.

\begin{figure}[!htbp]
    \centering
\includegraphics[width=\columnwidth]{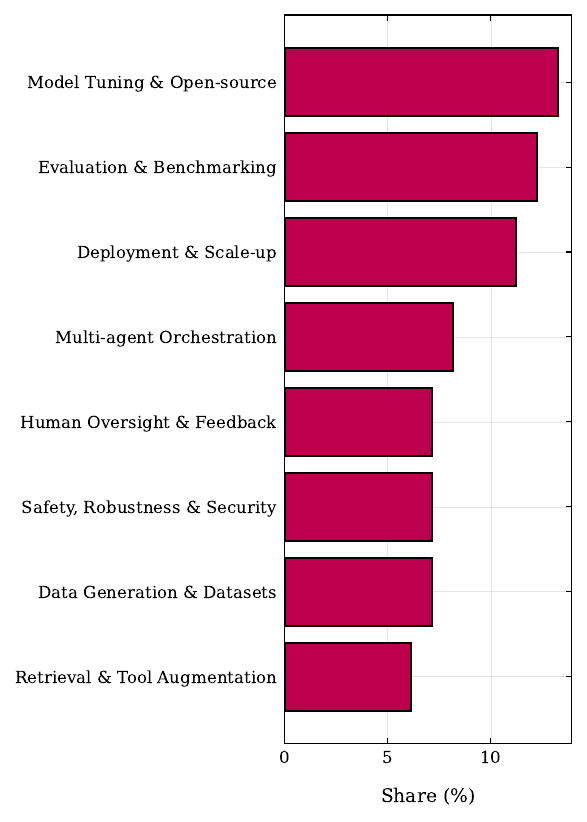}
    \caption{Top future work themes for \ac{FM}-based agent systems, measured by the relative share of papers mentioning each theme.}
    \label{fig:future-work}
\end{figure}

The future-work distribution in Figure~\ref{fig:future-work} is based on 98 total mentions and mirrors the limitation profile. Model tuning and open-source models (13, 13.3\%) leads the ranking; this theme covers domain-specific adaptation through fine-tuning, distillation, or open-weight alternatives, as well as calls for locally runnable models and public artifact releases. Evaluation and benchmarking (12, 12.2\%) addresses the external-validity gap through shared benchmarks, multi-criterion validation frameworks (combining expert evaluation, consistency analysis, and functional accuracy), user studies, and broader evaluation scopes. Deployment and scale-up (11, 11.2\%) calls for transitioning proofs-of-concept to real operational environments, broadening coverage across geographic regions and application domains, and enabling enterprise-level integration. Multi-agent orchestration (8, 8.2\%) indicates ongoing interest in advancing coordination strategies, domain-specialised agent roles, and adaptive orchestration mechanisms.

The remaining future-work themes address trustworthiness and infrastructure. Human oversight and feedback (7, 7.1\%) proposes structured approaches to balance agent autonomy with operational safety, including dynamic confidence thresholds, human review gates, and explicit feedback loops for in- and out-of-the-loop designs. Safety, robustness and security (7, 7.1\%) encompasses verification and observability mechanisms, robustness to adversarial prompt perturbations, privacy protections, and methods for interpretable and trustworthy decision-making. Data generation and datasets (7, 7.1\%) calls for physics-based or synthetic data generation, broader and more diverse benchmark datasets, and automated data ingestion pipelines to overcome the data scarcity identified above. Retrieval and tool augmentation (6, 6.1\%) advocates strengthening agent reasoning and correctness through domain-specific \ac{RAG}, integration of external computation tools and data sources, and structured context provision. As with limitations, mention frequency should not be equated with practical severity, since papers differ substantially in reporting depth and focus.

\section{Discussion}\label{sec:discussion}

Across the corpus, the evidence base for \ac{FM}-based technical agents in industrial and industrially relevant settings is characterised by prototype-oriented contributions and early-stage validation, while reports of operational deployments remain comparatively rare (RQ1). In terms of operational goals, most systems focus on user-assistance, monitoring, and resource-related process optimisation rather than in conventional production-control roles such as dispatching, scheduling, and planning (RQ2). Consistent with this purpose profile, the reported capability coverage emphasizes context handling, decision support, interaction, and communication with external tools and systems, whereas conventional multi-agent mechanisms such as negotiation are less frequently evidenced. Finally, the most frequently reported limitations and future-work directions point to practical deployment barriers and trustworthiness concerns: external validity and evaluation scope, reliability (including instability and hallucination), data constraints, latency and cost, and challenges at the interface between \acp{FM} and industrial automation infrastructures (RQ3).

\subsection{Adequacy of the working definition}
\label{subsec:discussion-definition}
The working definition introduced in Section~\ref{subsec:working-definition} requires (i)~encapsulation and specified objectives in an industrial context, (ii)~autonomous action beyond pure text generation, and (iii)~an \ac{FM} as a central component for decision-making and action selection. The corpus-level results allow a retrospective assessment of whether this three-part criterion adequately captures the systems reported in the literature.

The capability profile (Figure~\ref{fig:capabilities-comparison}) supports the definition's emphasis on encapsulation, decision-making, and autonomous action: encapsulation, decision-making capability, and reactivity are evidenced in over 90\% of the included systems, and external system communication, a proxy for actionable tool use beyond text output, exceeds 80\%. Context awareness is similarly prevalent, indicating that the included systems do not merely generate text but actively interpret situational information to select actions. These observations support the view that criterion~(ii) effectively separates \ac{FM}-based \ac{MAS} from purely generative \ac{LLM} applications.

The encapsulation and objective criterion~(i) is harder to validate empirically, because no publication in the corpus explicitly discusses encapsulation as a design property. However, encapsulation arguably arises \emph{by design} in \ac{FM}-based agents: external interaction is mediated exclusively through prompts, and the internal states, reasoning traces, and strategy of the model are not directly accessible to other system components at runtime. 
Combined with the prevalence of architectures in which individual agents are assigned distinct roles and objectives (Section~\ref{subsec:results-rq1}), this suggests that encapsulation is a de facto design principle of FM-based agents, not because authors deliberately engineer it, but because the prompt-based interaction paradigm inherently hides internal states from the external world.

A potential limitation of the definition concerns its scope: the strong focus on user assistance and monitoring in the corpus (Section~\ref{subsec:results-rq2}) raises the question of whether some assistive systems, for instance conversational knowledge-retrieval interfaces, satisfy the autonomy criterion or are better characterised as interactive tools. The conservative screening applied in this work mitigates this risk by requiring autonomous behaviour, but borderline cases remain. Future work could refine the autonomy threshold, for example by distinguishing degrees of autonomy along established taxonomies.

\subsection{Implications for research and practice}
\label{subsec:discussion-implications}
For industrial deployment, the observed maturity distribution and the limitation profile suggest that integration, latency, and safety assurance are currently key bottlenecks. In particular, the recurrent focus on tool use, executability, and interaction with existing automation layers indicates that many systems are not limited by the underlying \ac{FM} alone, but by the end-to-end engineering of reliable action execution and monitoring in production environments.

For evaluation, the predominance of early-stage evidence and the strong future-work emphasis on benchmarking and validation highlight the need for more consistent reporting and more deployment-near evaluation setups. In addition to task-level performance, evaluation should explicitly cover robustness, failure modes, and operational constraints (e.g., latency, compute costs, and data availability) that are decisive in industrial contexts.

For agent architecture design, the capability profile points to language-mediated interaction and knowledge-intensive reasoning as prominent value propositions, while negotiation- and synchronisation-heavy \acp{MAS} patterns appear less central in the current corpus. This suggests that practical designs often rely on tool-augmented, workflow-integrated architectures with explicit grounding, observability, and human oversight rather than on fully decentralised negotiation protocols.

\subsection{Threats to validity}
\label{subsec:discussion-validity}
This work is subject to selection effects and coverage bias induced by the chosen data sources, query formulation, and the focus on publications from 2020 onwards. In addition, the screening procedure is intentionally strict with respect to excluding purely conceptual work; consequently, the resulting corpus reflects research that reports implementations or case studies and may under-represent design-only contributions which, even though they lack empirical proof, still might provide valuable insights into existing capabilities.

Two selection effects at opposite ends of the \ac{TRL} scale may influence the maturity distribution reported in Section~\ref{subsec:results-rq1}. At the lower end, contributions at \ac{TRL}~1--3 are likely under-represented because the inclusion criteria (Section~\ref{subsec:screening}) require a concrete application, implementation, or case study; purely conceptual work, position papers, and early theoretical proposals are not considered. Early-stage research on \ac{FM}-based agent concepts that has not yet been accompanied by an implementation is therefore not reflected in the corpus, even if such work exists in the literature. At the upper end, systems operating at higher readiness levels in industrial practice may not always be reported in scientific publications, for instance due to intellectual property concerns, confidentiality agreements, or weaker publishing incentives for industrial practitioners once a system has left the prototype stage. The \ac{TRL} distribution can therefore be understood as reflecting the maturity landscape as visible in the scientific literature under the applied inclusion criteria, which should be kept in mind when interpreting the observed shares at the extremes of the scale.

Coding ambiguity is a second threat to validity. The analysis is generally performed at paper level and relies on explicit evidence in publications. One exception is encapsulation, which is set to \emph{supported} for all included systems by design rather than by textual evidence (see Section~\ref{subsec:discussion-definition}). If a paper omits architectural detail, a capability or purpose may be present in the implemented system but not evidenced for coding. Conversely, some categories may be described as intended functionality without strong empirical validation. These issues are particularly relevant for maturity (\acs{TRL}) assessments, which are often inferred from evaluation descriptions rather than stated explicitly. A related concern refers to the coding of real-time capability: several publications declare their system or individual actions as ``real-time'' without explicitly specifying the temporal resolution or response-time guarantees. In some cases, real-time capability is attributable to specific sub-components (e.g., sensor-based monitoring modules) rather than to the \ac{FM}-based agent itself. These claims were accepted at face value during coding, which may lead to an overestimation of real-time capability in the corpus. This coding ambiguity reflects a broader validity challenge in empirical synthesis: conclusions depend on the completeness, operational clarity, and triangulation of the reported evidence, while omitted detail may weaken the basis for stronger inferences~\cite{Yin2013Evaluation}.

The descriptive comparison against the baseline survey~\cite{reinpold2025agentsDTcomparison} is limited by differences in corpus composition, inclusion criteria, and reporting depth. The baseline covers industrial software agents and \acp{DT} broadly, whereas this work focuses specifically on \ac{FM}-based technical agents. In addition, differences in reporting depth and evidence operationalisation can affect whether a capability or purpose is counted as supported. Although the same purpose--capability framing is used, observed differences cannot solely be attributed to the \ac{FM} integration but might also stem from corpus-dependent coverage deviations.

Finally, the quantitative synthesis is largely based on mention frequencies and paper-level counts. Mention frequency does not necessarily correspond to practical severity or importance: papers differ in reporting depth and focus, and, for example, a limitation theme can be decisive for deployment even if it is mentioned infrequently.

\subsection{Outlook}
\label{subsec:discussion-outlook}
Based on the consolidated limitation and future-work themes, immediate next steps should include (i) more deployment-near evaluations and longitudinal studies that move beyond prototype evidence, (ii) shared evaluation protocols and benchmarking frameworks, and reporting templates for \ac{FM}-based agent systems in industrial contexts, (iii) robust tool and digital-infrastructure integration patterns (including monitoring, debugging, and auditability), and (iv) safety- and security-oriented engineering measures (e.g., fallback strategies, constraint enforcement, and governance processes). In addition, domain-relevant datasets remain important prerequisites for comparing approaches and for improving external validity across application areas. Looking at the distribution of application domains, logistics and supply chain appears to have been studied only rarely, which is another avenue of potential future work.

Beyond these consolidation-oriented directions, the results show a particularly interesting dynamic around diagnosis and fault management. As noted in Section~\ref{subsec:results-rq2}, this purpose is retained at near-baseline levels (24 assignments, 10.4\%; baseline 13.6\%; $-3.1$\,pp), unlike the sharper decline observed for planning, scheduling, and dispatching. This persistence is consistent with the capability profile of \ac{FM}-based agents: they can process and interpret heterogeneous failure symptoms, parse unstructured maintenance logs and operator reports, and leverage patterns learned during training about system behaviour and failure modes to support root-cause identification and the determination of adequate corrective actions. The combination of context awareness, dealing with uncertainty, and learning, which show the largest positive deviations relative to the baseline, aligns conceptually with diagnostic tasks. At the same time, the limitations identified in Section~\ref{subsec:results-rq3}, hallucination, output instability, and inference latency, remain particularly consequential for this purpose, since diagnostic tasks in industrial settings can be safety-relevant and time-critical. Translating the observed potential into reliable deployments therefore represents a promising direction for future work.


A further research opportunity concerns the systematic analysis of task decomposition in hybrid agent architectures. The capability profile shows that \ac{FM}-based agents excel in language-mediated interaction, context interpretation, and reasoning under uncertainty, yet conventional algorithmic methods remain superior for well-defined computational tasks such as scheduling and path planning. A principled investigation into which sub-tasks genuinely benefit from \ac{FM} reasoning and which are better delegated to established domain algorithms, e.g.\ based on negotiations and established protocols, would help practitioners design more efficient architectures that combine the strengths of both paradigms. Such an analysis could build on the purpose and capability dimensions introduced in this work and extend them with a task-level granularity that distinguishes generative, interpretive, and computational sub-tasks within a single agent workflow. Moreover, integrating classical and \ac{FM}-based agents with the \ac{DT} paradigm provides another possibility for investigation. DTs, that for example exhibit capabilities like simulation and prediction, can further enhance the spectrum of possible applications. 

\section{Conclusion}\label{sec:conclusion}
This work provides a systematic literature review of \ac{FM}-based technical agents in industrial and industrially relevant contexts and synthesizes a corpus of $N=88$ publications using a transparent, evidence-linked coding scheme. The results indicate that the reported systems are predominantly evaluated at prototype and early validation stages, with comparatively few publications providing deployment-oriented evidence (RQ1). Across the corpus, operational goals are most frequently positioned in assistive, monitoring, and resource-related process optimisation-centric roles, and the reported capability profile emphasises interaction, context handling, decision support, and tool-mediated communication, while conventional negotiation-heavy \acp{MAS} mechanisms are less frequently evidenced (RQ2). The most widely reported limitations and future work directions highlight external-validity gaps, reliability concerns, data constraints, and practical deployment barriers related to latency, cost, and system integration (RQ3). Overall, the proposed coding representation and comparative framing can serve as a reusable basis for future surveys and for more standardised evaluation and reporting of industrial \ac{FM}-based agent systems.

\backmatter

\bmhead{Acknowledgements}
This work was carried out within the VDI/VDE-GMA Technical Committee~3.35 on Industrial Agents. The authors thank the committee members for the valuable discussions and contributions that shaped this survey.

\section*{Declarations}

\begin{itemize}
\item \textbf{Funding:} Not applicable.
\item \textbf{Conflict of interest:} The authors declare no competing interests.
\item \textbf{Ethics approval:} Not applicable.
\item \textbf{Data availability:} Not applicable.
\item \textbf{Author contributions:} The first three authors are listed in order of contribution; the remaining authors are listed alphabetically by surname. V.H.: Conceptualization, Methodology, Formal analysis, Investigation, Data curation, Writing -- original draft, Writing -- review \& editing, Visualization. F.G.: Conceptualization, Methodology, Formal analysis, Writing -- original draft, Writing -- review \& editing, Supervision. D.K.: Data curation, Formal analysis, Writing -- review \& editing. A.A., B.H., C.L., F.M., F.O., L.C., L.V., M.O., M.T., N.A.T., P.K., T.S., Y.X.: Formal analysis, Writing -- review \& editing.
\end{itemize}

\bibliography{references}

\end{document}